\documentclass{llncs}
\usepackage{makeidx}  % allows for indexgeneration
\usepackage{etoolbox} % for "\patchcmd" macro
\usepackage{graphicx}
\usepackage{multirow}
\usepackage{array}
\usepackage{adjustbox}
\usepackage{color, soul}
\usepackage{tabularx}
\usepackage[inkscapelatex=false]{svg}

\makeatletter
\renewcommand{\@makefnmark}{\hbox{}} % 저자명 옆 기호 없애기
\renewcommand{\@makefntext}[1]{\noindent\footnotesize#1} % 각주에서 기호 없애고 글자 크기 줄이기
\patchcmd{\ps@headings}{\rlap{\thepage}}{}{}{}
\patchcmd{\ps@headings}{\llap{\thepage}}{}{}{}
\makeatother
\begin{document}
%
%\frontmatter          % for the preliminaries
%
%\pagestyle{headings}  % switches on printing of running heads
%\addtocmark{Hamiltonian Mechanics} % additional mark in the TOC

%\tableofcontents
%
\mainmatter              % start of the contributions
\title{Traversability-aware Consistent Situational Graphs for Indoor Localization and Mapping}
\titlerunning{T\textit{S-graphs}}  % abbreviated title (for running head)
\author{Jeewon Kim\inst{1} \and Minho Oh\inst{1} \and Hyun Myung\inst{1}\inst{,*} \\
\inst{*} Corresponding author
}
\authorrunning{Jeewon Kim et al.} % abbreviated author list (for running head)
\tocauthor{Jeewon Kim, Minho Oh, and Hyun Myung}
\institute{School of Electrical Engineering, KAIST (Korea Advanced Institute of Science and Technology), Daejeon, 34141, Republic of Korea,\\
\email{\{ddarong2000, minho.oh, hmyung\}@kaist.ac.kr} \\
\texttt{http://urobot.kaist.ac.kr}
\thanks{
This work was supported in part by the Technology Innovation Program(or Industrial Strategic Technology Development Program-Robot Industry Technology Development)(00427719, Dexterous and Agile Humanoid Robots for Industrial Applications) funded By the Ministry of Trade Industry \& Energy(MOTIE, Korea) and in part by the National Research Foundation of Korea (NRF) grant funded by the Korea government (MSIT)(RS-2024-00348461).
}
}
\maketitle

\begin{abstract}
Scene graphs enhance 3D mapping capabilities in robotics by understanding the relationships between different spatial elements, such as rooms and objects. 
Recent research extends scene graphs to hierarchical layers, adding and leveraging constraints across these levels. This approach is tightly integrated with pose-graph optimization, improving both localization and mapping accuracy simultaneously. 
However, when segmenting spatial characteristics, consistently recognizing rooms becomes challenging due to variations in viewpoints and limited field of view (FOV) of sensors. 
For example, existing real-time approaches often over-segment large rooms into smaller, non-functional spaces that are not useful for localization and mapping due to the time-dependent method. 
Conversely, their voxel-based room segmentation method often under-segment in complex cases like not fully enclosed 3D space that are non-traversable for ground robots or humans, leading to false constraints in pose-graph optimization.
We propose a traversability-aware room segmentation method that considers the interaction between robots and surroundings, with consistent feasibility of traversability information.
This enhances both the semantic coherence and computational efficiency of pose-graph optimization. 
Improved performance is demonstrated through the re-detection frequency of the same rooms in a dataset involving repeated traversals of the same space along the same path, as well as the optimization time consumption.

\end{abstract}
\section{Introduction}
Scene graphs offer a structured representation of 3D environments, enhancing semantic-rich mapping in robotics. Unlike traditional methods that focus solely on geometric data, this approach captures both objects and their spatial relationships, thereby improving navigation and understanding of surroundings for robots and humans alike. However, constructing accurate scene graphs heavily depends on precise robot localization, which relies on detailed elements like keyframes or point clouds. Recent improvements in SLAM methodologies conducted by Bavle, H. \textit{et al.} \cite{bavle2022situational,bavle2023s,millan2023better} include pose-graph optimization (PGO) to use geometric data across different layers for better localization and mapping.

However, the existing approaches for online room segmentation that utilize the Euclidean Signed Distance Field (ESDF) map~\cite{8202315} often produce results lacking an understanding of the interaction between the robot and its surroundings. 
Additionally, when entering a new space and segmenting spatial characteristics, consistently recognizing rooms becomes challenging due to voxel-based free space detection method and the time dependency of room segmentation method. For example, while some real-time approaches over-segment large rooms into smaller, non-functional spaces that are not useful for localization and mapping, others tend to under-segment in more complex cases, such as open spaces that are non-traversable for ground robots or humans but estimated as free space because the 3D space is not fully enclosed. This under-segmentation issue arises from the use of a voxel-based method, leading to false constraints in PGO.

To address these limitations, we propose a traversability-aware room segmentation method that aligns with conditions such as the field of view (FOV), making room segmentation using traversability information consistently feasible. 
This method demonstrates the following contributions:
\begin{itemize}
    \item[\textbullet] Our proposed traversability-aware free-space clustering method, which considers the traversability of the terrain, improves the consistency of room segmentation.
    \item[\textbullet] Consistent room segmentation by adding lower-layer data flushing rule, not using timer dependency.
    \item[\textbullet] The effectiveness of our method is shown by evaluating repetitive operations along the same path, specifically focusing on improvements in room re-detection rates and reductions in PGO processing time. 
\end{itemize}

\begin{figure}[h!]
\centering
\includegraphics[width=1.0\textwidth]{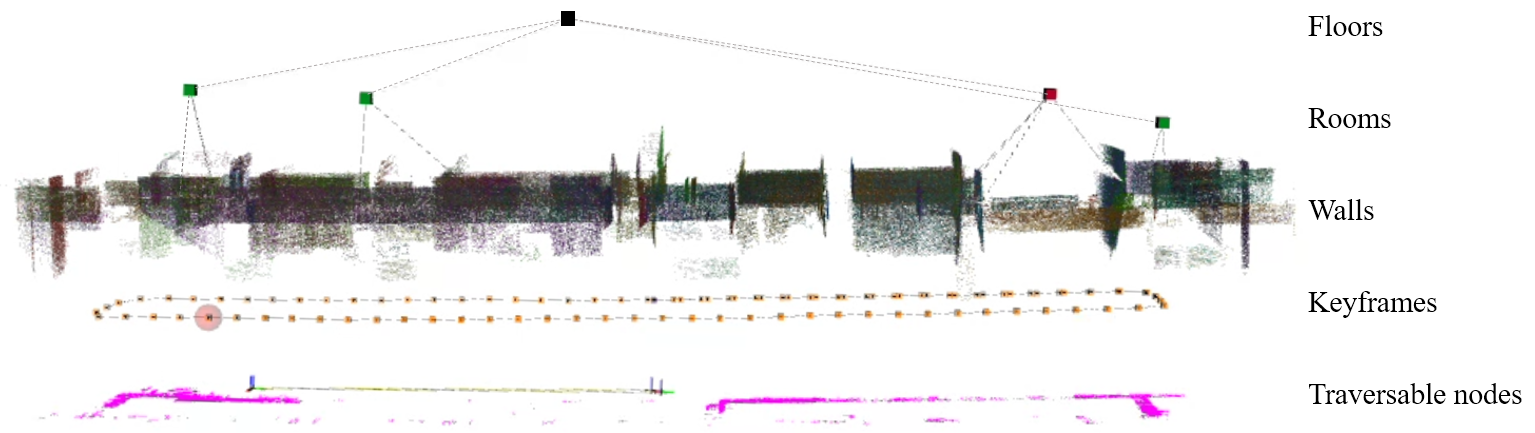} 
\caption{Our proposed hierarchical graph layers. From the bottom, the hierarchical graph consists of traversable nodes, keyframes, walls, rooms, and floors layers.}
\label{fig:main_figure}
\end{figure}
\section{Related Works}
\subsection{SLAM and Scene Graphs}
Many recent SLAM methods use geometric features from point clouds collected by LiDAR or depth camera~\cite{zhang2014loam,liosam}, which show robustness to complex environments and notable SLAM performance. These methods, however, are unable to handle tasks such as utilizing maps without semantic information or performing data association, such as loop closure, as they require processing entire point sets. Accordingly, semantic SLAM which assigns meanings to terrain-level or object-level points and enhances human-robot interactions is getting interests~\cite{shan2018lego,chen2019suma}. Still, challenges remain because the semantic segmentation accuracy suffers from sensor noise or localization error. Recent efforts aim to combine the strengths of geometric and semantic approaches by leveraging the geometric relationships between semantic elements \cite{9561933}. 

The concept of a scene graph for SLAM has emerged to capitalize on structural constraints across different layers, using a hierarchical approach. Studies like \cite{hughes2022hydra,werby2024hierarchical} employ visual sensors for object segmentation and map semantic labels, such as assigning a dining table to a kitchen, enhancing the level of detail in the graph. On the other hand, \cite{bavle2022situational,bavle2023s,millan2023better} only use point clouds to create layered scene graphs representing walls, rooms, and floors, with a foundational layer of keyframes. This hierarchical graph optimization results in more precise robot trajectories and better geometric reconstruction of indoor spaces. However, they were only tested on small and simple-shaped rooms, while they actually failed in larger or complex spaces like long corridors. Especially, in the most progressed work among them, Bavle, H. \textit{et al.}~\cite{millan2023better} introduces a Graph Neural Network (GNN) based approach for efficient semantic segmentation in SLAM systems, inferring relationships between wall surfaces to construct enriched graphs.

\subsection{Room Segmentation}
Accurate room segmentation is crucial because the room triggers the creation of new branch of the graph which affects the PGO process. In the approach described by S-graphs+~\cite{bavle2023s}, room segmentation is conducted in two main steps: ESDF map-based free-space clustering and room extraction using wall data. An ESDF map calculates the shortest distance from a free-space voxel to the nearest obstacle, typically a wall. By setting a distance threshold, edges and vertices close to walls or doorways are removed, allowing the graph to break apart near these boundaries, which helps in distinguishing between different rooms. The wall data from lower-layer and the free-space clusters are stacked for the fixed time intervals, and then the wall set lying near the clusters is chosen to define the rooms.

Furthermore, GNN-based approach~\cite{millan2023better} used a proximity graph and inferred relationships between walls for efficient room detection. The method identifies room structures quickly and improves detection time over S-graphs+.

\section{Traversability-aware Room Segmentation}
\begin{figure}[h]
    \centering
    \includegraphics[width=0.9\textwidth]{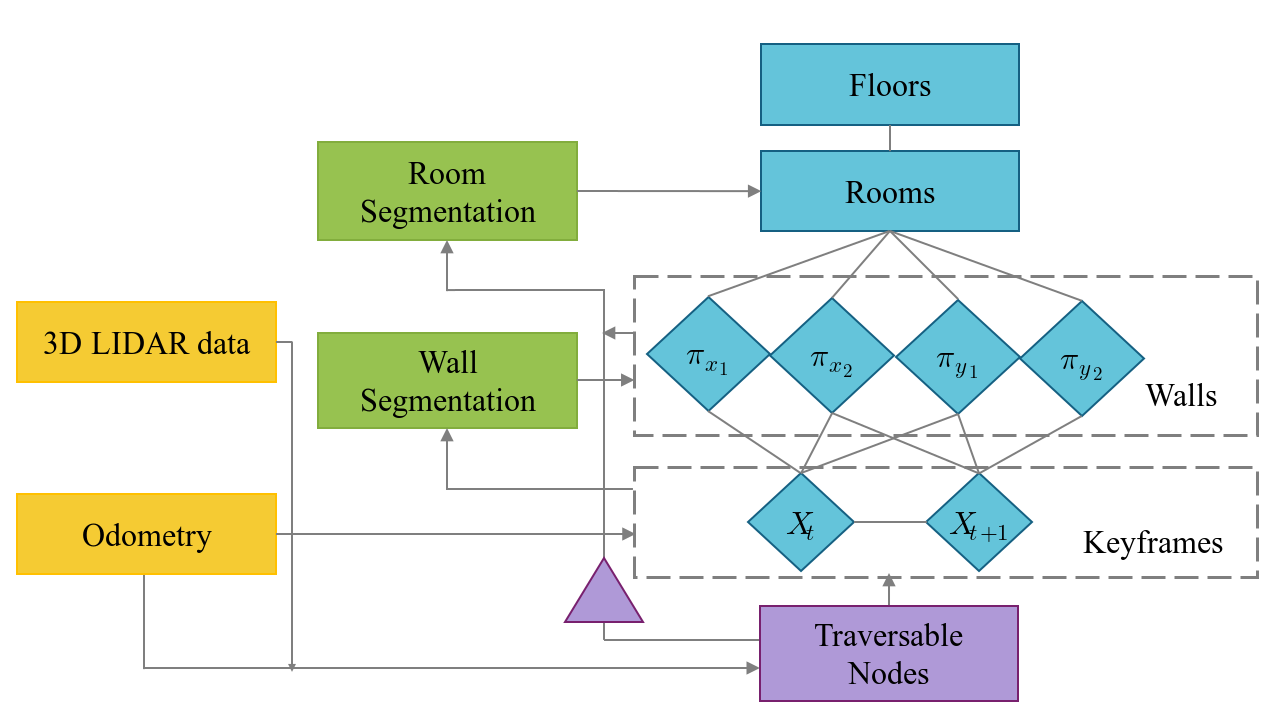}
\caption{Overview of our method. Inputs are the 3D LiDAR measurements, which is pre-filtered while calculating traversability, and robot odometry. Front-end is to extract wall planes, rooms, floors and traversable nodes. and the graphs are jointly optimized in the back-end. We added traversability-aware room segmentation strategy for consistency.}
\label{fig:overview}
\end{figure}

The overall architecture of our method is illustrated in Fig.~\ref{fig:overview}.
The situational graph layers are similar with \textit{S-graphs}+~\cite{bavle2023s}, but with additional traversable nodes layer and the strategy. We enhance room segmentation by implementing a traversability-aware strategy, which ensures the extraction of meaningful and consistent rooms. Here, meaningful room indicates traversable room for ground robots and humans, including interaction between them and terrain.

This method consists of three steps: extracting traversable ground nodes, clustering these nodes, and extracting rooms by utilizing the clustered traversable spaces.

\subsection{Traversable Ground Points Extraction}
In robotics, the demand for accurate detecting and representing the surrounding terrain is growing, especially for unmanned ground vehicles. Research in this area has primarily focused on identifying drivable regions and object identification~\cite{zermas2017fast,xue2021lidar}, odometry estimation~\cite{song23bigstep}, and global localization techniques~\cite{lim2024quatro++}.

We employ the ground segmentation method described in \cite{oh2022travel}, which identifies likely ground points based on their minimal height. To determine traversable areas, \cite{oh2022travel} calculates the local convexity and vertical alignment of each region and incorporates a Bayesian generalized kernel (BGK) to consider the surrounding points together. Additionally, we added an outlier points rejection step for better initial ground extraction and global update process to ensure global consistency in the map representation. Thereby, we obtained global consistent traversable terrain nodes.

\subsection{Free-space Clustering}
ESDF map-based free-space causes under-segmentation of room in complex indoor environments, such as corridors with central hollows as illustrated in Fig.~\ref{fig:example_open_space}, which appears frequently in many types of buildings. Segmentation inaccurately spans from wall to opposite one, as shown in Fig.~\ref{fig:qualitative_comparison}(a), instead of more logical boundaries such as from wall to handrail. Also, another baseline Bavle, H. \textit{et al.}~\cite{millan2023better} demonstrating faster and better room segmentation, showed the same limitation as the network was trained not considering the interaction between the robot and the surroundings.
%Also, ESDF map might differs with the height of the sensor, while the legged robot and human are mainly used indoors rather than aerial robots, so the height can't be changed easily. This brings different results from different platforms, disabling cooperation between data obtained by different platforms. Therefore, we used traversable ground points for free-space clustering. 

In this paper, we have developed a method that utilizes traversable ground points for free-space clustering. This process begins by constructing a node graph of ground region centers connected by edges, then selectively disconnected based on their proximity to occupied point clouds, with a threshold set at ${\lambda}_{th}$. Several clusters are extracted near the obstacles or doorways as a result, thereby we only consider a cluster that includes the robot's current position to avoid including areas outside doorways.

\subsection{Room Extraction Strategy}
Furthermore, while \textit{S-graphs}+ traditionally extracts rooms at every set time interval, which is effective in small or simple environments, it struggles with larger spaces or extended corridors.  In such scenarios, room definitions can become unclear, and inconsistent room identifications can occur at different times. To address the challenges in room segmentation, we introduce a novel strategy that utilizes changes in aisle width or direction as indicators. This method involves flushing the lower-layer data that are typically accumulated for room extraction. For instance, if a large bookshelf limits the examination of an entire room, the space between the wall and the bookshelf should be considered a distinct room. Doing so, it enhances the relevance and accuracy of room segmentation, ensuring that each defined room is meaningful and directly related to the navigational and operational context of the robot.

\section{Experimental Results}

In this section, we evaluate our proposed traversability-aware approach in terms of efficiency and consistency, comparing it with \textit{S-graphs}+~\cite{bavle2023s} and Bavle, H. \textit{et al.}~\cite{millan2023better}. 
We utilize the TIERS dataset~\cite{sier2023benchmark}, which includes a scenario where the starting point is revisited at the end. The odometry was calculated with scan-matching based SLAM algorithm~\cite{hdlslam}.
The experiment was run repetitively to demonstrate the efficiency and consistency of our approach. 
We conducted the experiment 10 times and averaged the results.

This experimental setup not only highlights the consistency and reliability of our approach but also demonstrates significant improvements in room segmentation and SLAM efficiency, achieving the objectives outlined in our methodological contributions.

\subsection{Consistent Room Segmentation}
\begin{figure}[hb!]
    \centering
    \includegraphics[width=\textwidth]{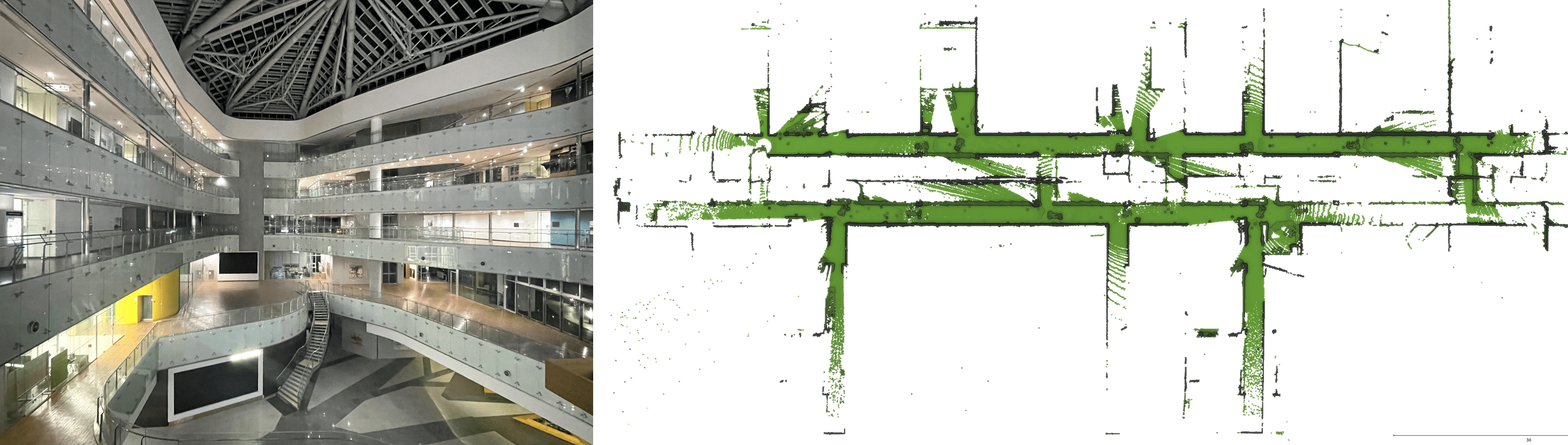}
    \caption{
        (L-R) Examples of complex corridor environments:
        (L) An open-corridor structure that shares 3D free space, where ground robots or humans cannot traverse. 
        (R) Bird's-eye view of the global point cloud map of the 3rd floor data from the Indoor sequence of the TIERS dataset, which traverses through open-corridor environments~\cite{sier2023benchmark}.
    }
    \label{fig:example_open_space}
    \vspace{2mm}
    \includegraphics[width=\textwidth]{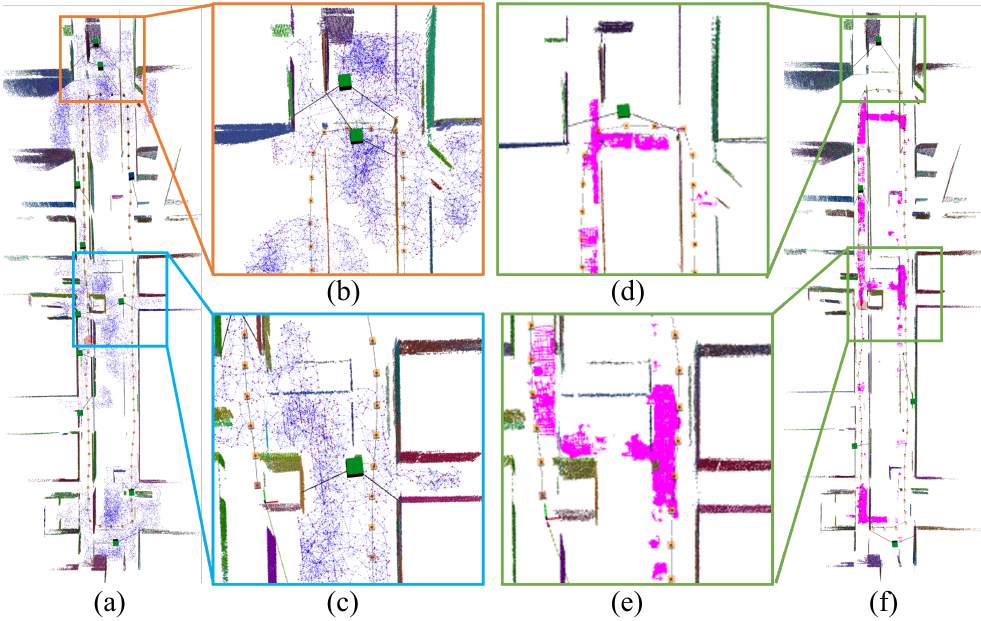}
    \caption{
        Qualitative results on the TIERS dataset~\cite{sier2023benchmark}:
        (a-c) The results of \textit{S-graphs}+~\cite{bavle2023s} demonstrate (b) over-segmentation and (c) under-segmentation in complex corridor environments due to the use of ESDF-based 3D free-space clustering (blue points).
        (d-f) The results of our proposed method show that by leveraging traversable nodes (magenta points) in free-space clustering, room segmentation is consistently maintained.
    } 
    \label{fig:qualitative_comparison}
\end{figure}

As shown in Fig.~\ref{fig:example_open_space}, it is an open space in three dimensions, so it is shared spatially, but it is actually a different layer and a different area, so it should be distinguished. 
However, in the case of existing baselines, when faced with that environment, spatial under-segmentation occurs, as shown in Fig.~\ref{fig:qualitative_comparison}(a).

In contrast, our method accounts for both robot and human traversability, resulting in reasonable free-space clustering as depicted in Fig.~\ref{fig:qualitative_comparison}(d-f), from the wall to the handrail.

The consistency of scene graphs was evaluated using the following metrics; frequency of room re-detection, and geometric precision metrics. First, we measured the ability to extract rooms and re-detect the same rooms in successive iterations. The re-detection ability was quantified by the ratio of the number of re-detected rooms to the number of rooms initially extracted in the first iteration, aiming for a high re-detection rate to confirm consistency.

Second, we assessed the precision of our segmentation with two further measures: Dice Coefficient Score (DCS) and distance of room center for the most overlapped rooms from the first and second traverse.
The Dice coefficient measures the similarity between two sets. For two square-shaped rooms \( A \) and \( B \), it is defined as:

\begin{equation}
{DCS}(A, B) = \frac{2 |A \cap B|}{|A| + |B|}
\end{equation}

Here, \( A \cap B \) is the intersection area, while \( |A| \) and \( |B| \) are the areas of room \( A \) and \( B \) respectively. DCS indicates strong overlap between the rooms segmented from the initial and second traverse in this paper, and the distance of room centers reflects minimal spatial deviation between the positions of the most overlapped room pairs. The average result values of the 10 iterative evaluations are shown in Table~\ref{tab:eval}.

% \begin{table}[t!]
% \centering
% \caption{Room segmentation consistency evaluation on repetitive run on the same place}
% \adjustbox{max width=\textwidth}{%
%     \begin{tabular}{|c|c|c|c|c|c|c|c|c|c|}
%         \hline
%         Metrics & \multicolumn{3}{c|}{\textit{S-graphs}+} & \multicolumn{3}{c|}{\textit{S-graphs}++} & \multicolumn{3}{|c|}{Ours} \\ \hline
%         Room category & Lap 1 & Lap 2 & Re-detected & Lap 1 & Lap 2 & Re-detected & Lap 1 & Lap 2 & Re-detected \\ \hline
%         Number of extracted rooms & 10.5 & 7.6 & 5.3 &  2.8 & 1.2 & 0 & 4.7 & 4.5 & 4.0 \\ \hline
%         Room re-detection frequency & \multicolumn{3}{c|} {50\%} & \multicolumn{3}{c|} {50\%} & \multicolumn{3}{|c|} {85\%}   \\ \hline
%         Dice coefficient score & \multicolumn{3}{c|} {61\%} & \multicolumn{3}{c|} {-} & \multicolumn{3}{|c|} {74\%}   \\ \hline
%         Room center distance (m) & \multicolumn{3}{c|} {0.97} & \multicolumn{3}{c|} {3.67} & \multicolumn{3}{|c|} {0.57}   \\ \hline
%     \end{tabular}
% }
% \label{tab:consistency}
% \end{table}

The baseline methods may produce slightly different predictions depending on how message-passing evolves across layers, and learned embeddings especially for GNN-based method ~\cite{millan2023better}. The variability in these predictions could also contribute to inconsistent room segmentation results. Whereas they extract over 10 rooms in the first traverse and re-detect only less than half among them, our method extracts a smaller number of rooms but is meaningful with traversability information, so the re-detection frequency was significantly high.

Also, higher DCS and smaller room center distance than S-graphs+ demonstrated consistent room segmentation in our method. In the case of  ~\cite{millan2023better}, room width was not provided in the room information but only room center position, so the DCS calculation requires an additional process. However, we considered it unnecessary to calculate the DCS for this method due to the inconsistency shown with $f_{re}$. 

\subsection{Efficient and Robust Pose-graph Optimization}

To demonstrate that the proposed method improves not only the pose estimate performance but also the efficiency, we evaluate the absolute trajectory error (ATE) of the distance between the start and end point position on \textit{Indoor} sequence of the Tiers dataset.

The ATE of the distance between the start and end point position was used to evaluate the navigational accuracy of SLAM. As the dataset returns to the start point at the end, the ground truth is zero, and the smaller value would indicates better performance.
We measured and compared the time efficiency of the hierarchical PGO process across the data to show the enhancements in SLAM operational speed facilitated by our method. We tracked each time consumption spent for PGO performed while the same rooms were detected and calculated the average value and overall summation during the whole data time.

% \begin{table}[h!]
% \centering
% \caption{PGO efficiency}
% \begin{tabular}{|c|c|c|c|}
% \hline
% Metrics                 & \textit{S-graphs}+   & \textit{S-graphs}++    & Ours \\ \hline
% ATE (m)                 & ~0.24                         & ~0.48                                      & ~0.22 \\ \hline
% $T_{PGO}$ (s)    & ~1.65                         & 2.66                                      & ~1.65 \\ \hline
% \end{tabular}
% \label{tab:time}
% \end{table}

\begin{table}[ht!]
    \vspace{2mm}
    \caption{
    Quantitative comparison for room segmentation consistency and pose-graph optimization efficiency. Room segmentation consistency was compared in terms of the frequency of room re-detection ($f_{re}$) of two iterations in TIERS LiDAR dataset, calculated by $N_{re}$ over $N^{1st}$, where $N^{1st}$ and $N^{2nd}$ indicate the numbers of rooms extracted from the first and second iteration and $N_re$ indicates the number of rooms detected again that are the same as the rooms in the first iteration among the rooms in the second iteration. Also, room consistency was compared with the similarity between the pair of rooms segmented from two successive traverses, using Dice Coefficient Score (DCS) and distance of room center ($d_{center}$) measured in [$m$]. Also, pose-graph optimization efficiency was compared with ATE measured in [$m$] and PGO computation time during the whole process ($T_{PGO}$) measured in [$s$].
    }
    % \vspace{-1mm}
    \setlength{\tabcolsep}{4.7pt}
    \setlength\extrarowheight{1pt}
\begin{tabularx}{\textwidth}{c||cccccc||cc}
\hline
\multirow{2}{*}{Metrics}  & \multicolumn{6}{c||}{Room Segmentation Consistency}                                                                                                                   & \multicolumn{2}{c}{PGO Efficiency}  \\ \cline{2-9} 
           & \multicolumn{1}{c|}{$N^{1st}$} & \multicolumn{1}{c|}{$N^{2nd}$} & \multicolumn{1}{c|}{$N_{re}$} & \multicolumn{1}{c||}{$f_{re}$ $\uparrow$} & \multicolumn{1}{c|}{DCS $\uparrow$} &  \multicolumn{1}{c||}{$d_{center}$ $\downarrow$ }  & \multicolumn{1}{c|}{ATE $\downarrow$} & $T_{PGO}$ $\downarrow$ \\ \hline
S-graphs+~\cite{bavle2023s}  & \multicolumn{1}{c|}{10.5}      & \multicolumn{1}{c|}{7.6}      & \multicolumn{1}{c|}{5.3}    & \multicolumn{1}{c||}{0.50}    & \multicolumn{1}{c|}{0.61}    &   \multicolumn{1}{c||}{0.97}   & \multicolumn{1}{c|}{0.24}    &   {1.68}   \\
Bavle, H. \textit{et al.}~\cite{millan2023better} & \multicolumn{1}{c|}{9.25}      & \multicolumn{1}{c|}{7.5}      & \multicolumn{1}{c|}{1.5}    & \multicolumn{1}{c||}{0.16}    & \multicolumn{1}{c|}{-}    &    \multicolumn{1}{c||}{3.67}    & \multicolumn{1}{c|}{0.48}    &   2.66   \\
Ours       & \multicolumn{1}{c|}{4.7}      & \multicolumn{1}{c|}{4.5}      & \multicolumn{1}{c|}{4.0}    & \multicolumn{1}{c||}{\textbf{0.85}}    & \multicolumn{1}{c|}{\textbf{0.74}}    &   \multicolumn{1}{c||}{\textbf{0.57}}  & \multicolumn{1}{c|}{\textbf{0.22}}    &    \textbf{1.62}  \\ \hline
\end{tabularx}
\label{tab:eval}
\end{table}

Table \ref{tab:eval} tells GNN-based method~\cite{millan2023better} brings even lower re-detection rate of the rooms than non-learning based method, as a result strongly depends on the prior learned model and not robust to various situations. This over-segmentation leads to high ATE and increased PGO time consumption.

Our method showed the most promising performance in ATE and PGO time consumption compared to \textit{S-graphs}+. \textit{S-graphs}+ tends to extract the duplicated room in successive timings, giving ambiguity and additional useless edges to the situational graphs, which brings a negative effect on the PGO and also takes more time. Our method addressed this by extracting room with one characteristic just once, improving efficiency of PGO.

\section{Conclusion}
In conclusion, our proposed traversability-aware strategy substantially improves both the semantic meaningfulness including robot-terrain interaction information, and consistency of the situational graphs. The results demonstrate that our method achieves a higher frequency of room re-detection and improved DCS compared to \textit{S-graphs}+ and \cite{millan2023better}. This suggests a more reliable and consistent construction of situational graphs, and accordingly, this brings accurate SLAM performance and reduced time consumption for PGO. These results confirm the effectiveness of our methodology in advancing scene graph-based SLAM systems.

For future work, we plan to explore how free-space clustering with an ESDF map and traversable nodes may vary depending on sensor orientation and height. We also aim to assess segmentation consistency across different robot platforms to offer cooperative scene graph construction in multi-platform environments.

% \section{Acknowledgement}
% This work was supported by the Technology Innovation Program(or Industrial Strategic Technology Development Program-Robot Industry Technology Development)(00427719, Dexterous and Agile Humanoid Robots for Industrial Applications) funded By the Ministry of  Trade Industry & Energy(MOTIE, Korea). The students are supported by BK21 FOUR.

\bibliographystyle{splncs03} % llncs와 호환되는 BibTeX 스타일
\bibliography{references} % reference.bib 파일을 참조

\end{document}